\documentclass{article}

\usepackage[final]{neurips_2021}

\usepackage[utf8]{inputenc} 
\usepackage[T1]{fontenc}    
\usepackage{hyperref}       
\usepackage{url}            
\usepackage{booktabs}       
\usepackage{amsmath}       
\usepackage{amsfonts}       
\usepackage{nicefrac}       
\usepackage{graphicx,wrapfig}
\usepackage{microtype}      
\usepackage{xcolor}         
\usepackage{subcaption}
\graphicspath{Figures/}

\title{Insight Miner: A Time Series Analysis Dataset for Cross-Domain Alignment with Natural Language}

\author{%
    Yunkai Zhang \thanks{Work done during internship at Google X - Mineral.ai.}\\
  UC Berkeley\\
  \texttt{yunkai\_zhang@berkeley.edu} \\
  \And
    Yawen Zhang \\
  Mineral \\
  \texttt{yawenz@mineral.ai} \\
  \And
  Ming Zheng \\
  Mineral \\
  \texttt{zhengming@mineral.ai} \\
  \And
    Kezhen Chen \\
  Mineral \\
  \texttt{ kezhenchen@mineral.ai} \\
  \And
  Chongyang Gao\footnotemark[1]\\
  Northwestern University \\
  \texttt{Chongyanggao2026@u.northwestern.edu} \\
  \And
  Ruian Ge\\
  UC Berkeley\\
  \texttt{ruian\_ge@berkeley.edu} \\
  \And
  Siyuan Teng\\
  UC Berkeley\\
  \texttt{siyuan\_teng@berkeley.edu} \\
  \And
  Amine Jelloul\\
  UC Berkeley\\
  \texttt{amine\_jelloul@berkeley.edu} \\
  \And
  Jinmeng Rao \\
  Mineral \\
  \texttt{jinmengrao@mineral.ai} \\
  \And
  Xiaoyuan Guo \\
  Mineral \\
  \texttt{xiaoyuanguo@mineral.ai} \\
  \And
  Chiang-Wei Fang \\
  UC Berkeley\\
  \texttt{chiangwei\_fang@berkeley.edu} \\
  \And
  Zeyu Zheng \\
  UC Berkeley\\
  \texttt{zyzheng@berkeley.edu} \\
  \And
  Jie Yang \\
  Mineral \\
  \texttt{yangjie@mineral.ai} \\
}

\begin{document}

\maketitle

\begin{abstract}
  Time-series data is critical across many scientific and industrial domains, including environmental analysis, agriculture, transportation, and finance. However, mining insights from this data typically requires deep domain expertise, a process that is both time-consuming and labor-intensive. In this paper, we propose \textbf{Insight Miner}, a large-scale multimodal model (LMM) designed to generate high-quality, comprehensive time-series descriptions enriched with domain-specific knowledge. To facilitate this, we introduce \textbf{TS-Insights}\footnote{Available at \href{https://huggingface.co/datasets/zhykoties/time-series-language-alignment}{https://huggingface.co/datasets/zhykoties/time-series-language-alignment}.}, the first general-domain dataset for time series and language alignment. TS-Insights contains 100k time-series windows sampled from 20 forecasting datasets. We construct this dataset using a novel \textbf{agentic workflow}, where we use statistical tools to extract features from raw time series before synthesizing them into coherent trend descriptions with GPT-4. Following instruction tuning on TS-Insights, Insight Miner outperforms state-of-the-art multimodal models, such as LLaVA \citep{liu2023llava} and GPT-4, in generating time-series descriptions and insights. Our findings suggest a promising direction for leveraging LMMs in time series analysis, and serve as a foundational step toward enabling LLMs to interpret time series as a native input modality.
\end{abstract}

\section{Introduction}

Time series data is fundamental to a wide range of domains. Traditionally, researchers have relied on statistical tools to analyze and interpret this data. Methods such as Autoregressive Integrated Moving Average (ARIMA)~\cite{box2015time}, Seasonal-Trend Decomposition using LoESS (STL)~\cite{cleveland1990stl}, and state space models~\cite{hamilton1994state} have long been the standard for forecasting, detecting seasonality, and understanding underlying trends. These techniques are particularly prevalent in fields like economics, meteorology, and transportation, providing essential interpretability for complex temporal data.

Recently, many studies have explored the usage of LLMs for time-series tasks. Several studies have leveraged pretrained LMs (e.g., GPT-2) for tasks such as forecasting, classification, and anomaly detection~\cite{zhou2023one,chang2023llm4ts}, achieving state-of-the-art performance and demonstrating the universality of pretrained representations. Other works have designed structured prompts to enable zero-shot or few-shot inference~\cite{xue2022promptcast,yu2023temporal}. However, these approaches primarily focus on tasks where the output is numerical (scalars or future time steps). Directly training LLMs to perform traditional numerical tasks does not inherently enable them to generate natural language insights or explain the data.

On the other hand, the emergence of Large Multimodal Models (LMMs), such as LLaVA~\cite{li2023llava}, has inspired new approaches to align domain-specific data with language models.A notable example is FinVis-GPT~\cite{wang2023finvis}, which builds upon LLaVA to generate a financial task-oriented dataset for alignment and instruction tuning. While FinVis-GPT demonstrates the feasibility of LMMs in analyzing financial charts, our work aims to generalize this success beyond a specific domain. We focus on constructing a time series analysis dataset for LMMs. To the best of our knowledge, no such dataset currently exists for aligning general time-series data with comprehensive textual descriptions.

In summary, our main contributions are two-fold: 1) We present a time series analysis dataset that enables LLMs to generate faithful and descriptive time series insights; and 2) We provide the first large-scale repository aligning time-series data into the language embedding space, paving the way for future research on LMMs for time-series data analysis.

\section{TS-Insights Dataset}
 To our knowledge, there are no existing large-scale datasets of time series and language description pairs, let alone for time series analysis. To bridge this gap, we design and generate the \textbf{TS-Insights Dataset}, the first dataset specifically curated for general time series analysis alignment.

Formally, given $N$ time series datasets $\{\mathcal{D}_i\}_{i=1}^N$, where each dataset $\mathcal{D}_i$ has $T_i$ total time steps and $M_i$ features, i.e., $D_i = \{X_j\}_{j=1}^{T_i}$ and $X_j \in \mathbb{R}^{M_i}$, the goal is to generate a question-answer pair for each time series window $W_k \in \mathbb{R}^{m_k \times \tau_k}$ randomly sampled from the $N$ datasets, where $\tau_k$ represents the number of time steps and $m_k$ represents the number of features, which are both randomly subsampled from the chosen dataset.\footnote{To generate the current dataset, $\tau_k$ is randomly sampled from [30, 500].} Each training sample consists of a time series window $W_k$, a question $L^Q_{k}$, and an answer $L^A_{k}$. Using $(W_k, L^Q_k, L^A_k)$, we create a single-round instruction-following example \citep{liu2023llava}:
\begin{equation}
\textbf{Human: } W_k\backslash\text{n }L^Q_k <\text{STOP}>\backslash \text{n    
   }\textbf{Assistant: } L^A_k<\text{STOP}>\backslash \text{n}.
\end{equation}

To generate such datasets for modalities such as images \citep{liu2023llava} or biomedical images \citep{li2023llavamed}, the common practice is to prompt a language-only model (e.g., GPT-4). For example, LLaVA \citep{liu2023llava} asks GPT-4 to generate multi-turn conversations based on image captions and object bounding boxes. However, the time series modality presents unique challenges: 1) there are no original captions available for a time series window, 2) existing tools cannot readily convert a time series segment into an input format that is suitable for language-only GPTs, and 3) the semantic meanings of time series windows are more difficult to be described in natural languages.

\begin{wrapfigure}{r}{7cm}
	\centering
    \includegraphics[scale=0.5]{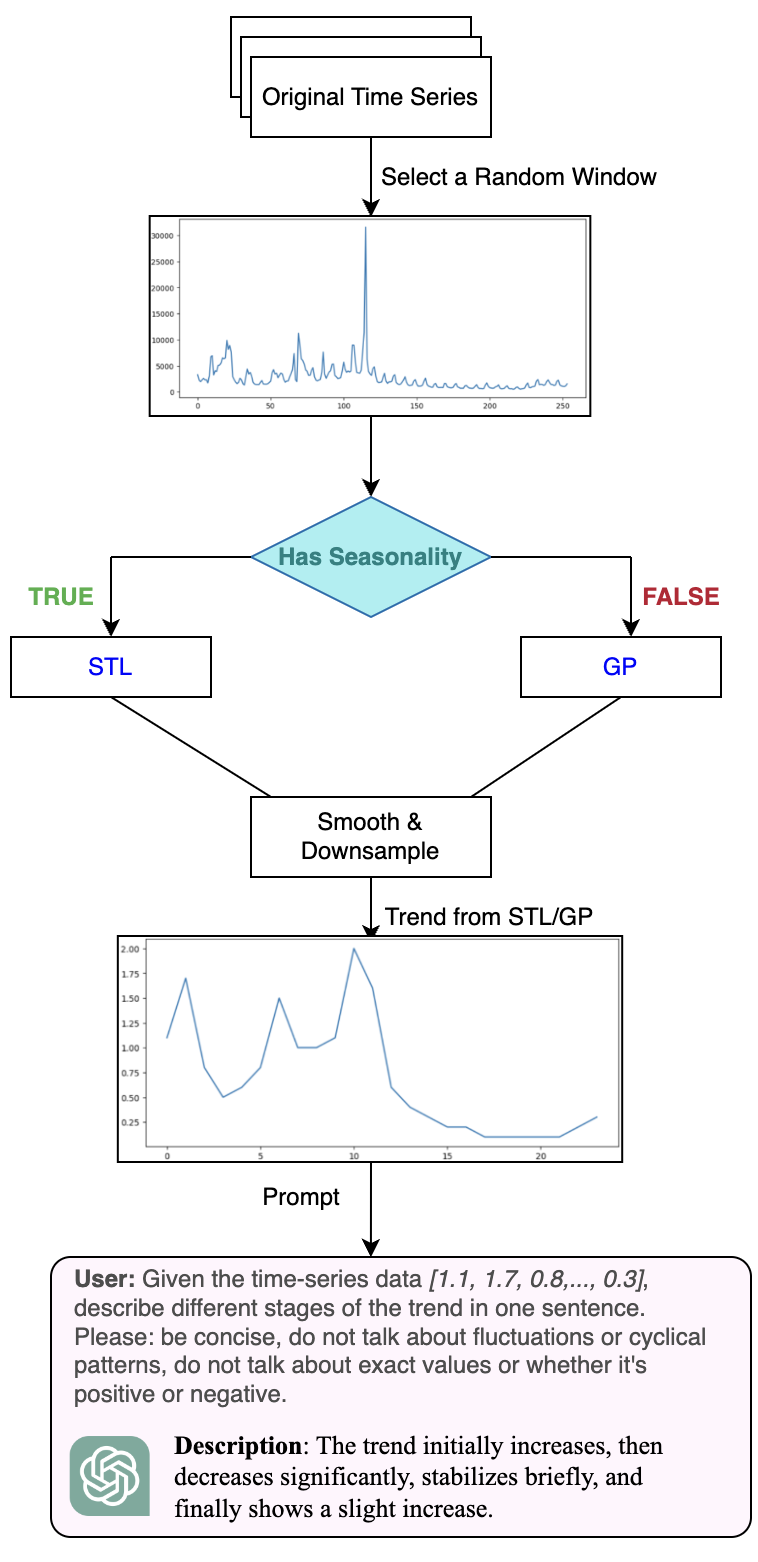}
    \caption{Trend dataset generation workflow using statistical tools.}
\label{fig: workflow}
\end{wrapfigure}

To address the third challenge, we focus on time series windows that contain a single feature, i.e., $W_k \in \mathbb{R}^{1 \times \tau_k}$, and following traditional time series analysis \citep{Brockwell1991}, we generate descriptions based on the trend, the seasonality, and the residuals that are present in the window. A naive solution is to feed in the raw time series as a vector when prompting GPT-4, e.g., "Given the time series [0.52, 0.98, 0.95, 0.91, 1.24, ..., 1.32], generate a description about its trend, seasonality, and volatility." However, we found that GPT-4 fails to accurately extract each component from the raw vector.\footnote{See Appendix \ref{sec: case-studies} for failure cases.} 

Instead, we implement a tool-use pipeline where we leverage the Seasonal-Trend Decomposition (STL) algorithm to mathematically decompose the original time series into its constituent trend, seasonality, and residual components. We then generate a description based only on one component at a time. As a proof of concept, we focus on the trend description in the current version of this paper.

\subsection{Trend Generation Workflow} \label{sec: workflow}
To generate the trend description for a given time series window $W_k \in \mathbb{R}^{1 \times \tau_k}$, we first apply an STL decomposition to extract the trend
\begin{equation}\label{eq: STL-overall}
    W_k = \mathcal{T}_k + \mathcal{S}_k + \mathcal{R}_k,
\end{equation}
where $\mathcal{T}_k$, $\mathcal{S}_k$, and $\mathcal{R}_k$ denote the extracted trend, seasonality, and residual components, respectively. We denote the value at each time step of the extracted trend as $\mathcal{T}_k = (\hat{y}_1, \hat{y}_2, \cdots, \hat{y}_{\tau_k})$.

In some cases, $W_k$ might not exhibit any seasonalities. In such cases, we fit a Gaussian Process (GP) to the $\tau_k$ time steps in the window. Let $W_k = (y_1, y_2, \cdots, y_{\tau_k})$, where $y_i$ represents the value at time step $i$. $W_k$ is modeled by a standard zero-mean GP, whose covariance structure is defined by a kernel $K(., .)$. Here, the kernel used is a combination of a Radial Basis Function (RBF) kernel to model the dependency among different time steps and a white-noise kernel to model the observational noise. That is, $W_k \sim GP(\mu(x), K(x, x'))$, where $\mu(x)=0$, $K(x, x')=RBF(x, x')+\sigma^2_e\delta_{x,x'}$, $RBF(x, x')=\sigma^2_r exp(-\frac{(x - x')^2}{2\gamma})$ and $\delta_{x,x'}$ is the Kronecker delta. The parameters $\sigma^2_r$, $l$ and $\sigma^2_e$ are estimated from the data by maximizing the likelihood. We then compute the posterior mean of the Gaussian Process regression at the respective time steps to get $\mathcal{T}_k = (\hat{y}_1, \hat{y}_2, \cdots, \hat{y}_{\tau_k})$ as the extracted trend.

We then apply a Gaussian kernel $\mathcal{F}_k = [\mathcal{F}_1, \mathcal{F}_2, \cdots, \mathcal{F}_{w_k}]$, where $w_k$ is a hyperparameter for the kernel size, to further smooth out the trend, and followed by downsampling with stride size $s_k$\footnote{We choose stride size $s_k$ so that $\tau_k // s_k = 25$.}: 
\begin{equation}\label{eq: smooth-and-downsample}
\tilde{y}_i = \sum_{j=-w_k // 2}^{w_k // 2} \hat{y}_{s_k\cdot i-j} \cdot \mathcal{F}_{w_k // 2 + j}
\end{equation}
for $i = 1, 2, \cdots, \tau_k // s_k$.

Finally, we round each entry of $(\tilde{y}_1, \tilde{y}_2, \cdots, \tilde{y}_{\tau_k // s_k})$ to one decimal place and feed it to GPT-4. As such, one data sample pair consists of the original time series window $W_k$ and the trend description generated by GPT-4. An overview of the workflow and the exact prompt we use is shown in Figure \ref{fig: workflow}.

\subsection{Trend Description Dataset}
Using the methodology described above, we generate 10,000 initial samples derived from 29 datasets in the Monash Time Series Forecasting Archive \citep{godahewa2021monash}. We reserve 11 additional datasets as a holdout set, which are only used for evaluation but not for training. The 29 datasets span a wide range of domains, including energy \citep{lai2018}, weather, traffic \citep{Jean-Michel_2019}, and healthcare \citep{covid}. Notably, we only sample windows from the train split of each dataset, defined to be the first $70\%$ of the time steps in temporal order. Some datasets contain multiple levels of seasonalities, e.g., daily and weekly. Under the original granularity, each window might not contain enough time steps to discern the higher level of seasonalities, since at least two full cycles are required to conclude there to be a seasonality. As trends should be described after seasonalities are removed, for each dataset, we also aggregate multiple time steps into one time step in order to introduce samples with more diversified patterns.

To further increase the number of training samples in a cost-efficient manner, for each GPT-4 labeled sample pair, we additionally apply nine different random augmentations to the original time series window $W_k$ such that the trend description is still applicable to the augmented samples. We then rephrase the original description generated by GPT-4 using GPT-3.5-turbo in order to increase the language diversity. Therefore, for each original sample, we now have nine augmented samples, resulting in 100k total training samples. A detailed list of test and holdout datasets, the number of samples we generate for each aggregated granularity level, as well as a list of augmentation methods can be found in Appendix \ref{sec: dataset-details}.

\section{Insight Miner}
We initialize our model using the pre-trained weights of LLaVA \citep{liu2023llava}, a state-of-the-art general-domain vision-language model, and continue finetuning the LLaVA weights to the time series domain. We use the same neural network architecture as LLaVA: we first convert the time series window into an image using a line plot, feed the image into the vision encoder, and then use a linear projection layer to map the vision output into the language embedding space. Finally, the language model takes in the projected image embeddings concatenated with the language instructions as the input and returns the language response. 

To align the time-series images with the LLM, we only finetune the linear projection layer, while keeping both the vision encoder and the language model frozen. For each training sample, we show the original time series to the model in the form of a line plot and the language instruction is to ask it to describe the trend, and the goal is to predict the description generated by the GPTs. The final model is named Insight Miner.

Note that the training cost of Insight Miner is relatively affordable as it was trained using 8 $\times$ A100 40GiB GPUs. Each epoch takes around an hour to train. Once the model finishes training, it can be easily deployed at a low inference cost.

\section{Experiments}\label{sec: experiments}
We conduct experiments to evaluate how well the trend dataset can enable large multimodal models to generate trend descriptions that are faithful to the original time series. More specifically, we sample 119 total windows for evaluation. Among these, 69 examples are from the test split (last 30\%) of the same datasets we used for training, and the other 50 examples are from the holdout datasets which are not used for training entirely. The models we include for comparison are: 
\begin{itemize}
    \item LLaVA \citep{liu2023llava}: using the checkpoint publicly available on \href{https://huggingface.co/liuhaotian/llava-v1-0719-336px-lora-merge-vicuna-13b-v1.3}{HuggingFace}.
    \item Vision (3 epochs): finetuned from the above LLaVA checkpoint for three epochs using the generated trend dataset. It takes in the original time series window plotted using the lineplot function in the Seaborn package.
    \item Vision (1 epoch): finetuned from the above LLaVA checkpoint for one epochs using the generated trend dataset.
    \item Engineering GPT: GPT-4 that takes in the extracted features as described in Section \ref{sec: workflow}.
\end{itemize}
Here, Vision (3 epochs) and Vision (1 epoch) are two versions of our Insight Miner trained using a different number of epochs. As we observed feeding the raw time series vector into GPT-4 leads to inferior descriptions compared to Engineering GPT, we do not include it for evaluation in this section, but it is included in the eight case studies shown in Appendix \ref{sec: case-studies}, along with the other four models.

\begin{wrapfigure}{r}{8cm}
    \includegraphics[scale=0.24]{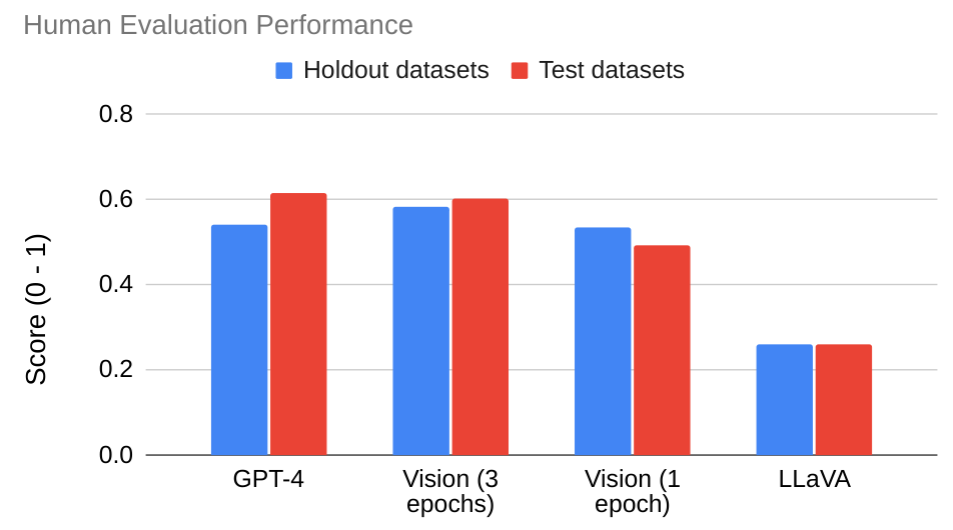}
    \caption{Description evaluation of different models by domain experts.}
    \label{fig: results}
\end{wrapfigure}

For each of the 119 samples, we generate one description using each of the above models, and ask three domain experts to manually score the descriptions generated. When presented to the domain expert, the descriptions from different models are shuffled in a random order for each sample. A score of 2 is given if the description matches the original time series, a score of 1 is given if the description is partially correct, and a score of zero is given if the description is not correct. We sum the scores from all human evaluators for all test (holdout) samples and normalize it to $0-1$ to produce the final score for each model. The results are summarized in Figure \ref{fig: results}.

As we see, both of our models, Vision (3 epochs) and Vision (1 epoch), significantly outperforms the original LLaVA model. Additionally, training for more epochs seems to lead to a better performance. In fact, using the vision encoder trained for three epochs can lead to a performance that is competitive to GPT-4, although the latter requires first preprocessing the time series using heuristics and statistical tools. Notably, Vision (3 epochs) outperforms GPT-4 on the holdout datasets. We hypothesize that this is because the holdout datasets contain more datasets with complicated seasonalities than the test datasets. Even though Engineering GPT-4 has access to the extracted features, it essentially still performs zero-shot inference. In comparison, our model is finetuned using the proposed TS-Insights dataset and can better leverage the abundance of labeled samples.

\section{Discussions}
This work presents the first large dataset with 100k training samples for general time series analysis in the form of time series and natural language pairs. We show that the proposed dataset can enable existing large multimodal models to align time series data with textual descriptions and perform detailed analysis.

In addition to the models evaluated in Section \ref{sec: experiments}, we also tried to use OneFitsAll \citep{zhou2023one} as the time-series encoder to replace the vision encoder in LLaVA. Our initial attempt shows that using a time-series encoder causes the model to fail to generate coherent descriptions for most samples, which is likely due to that unlike the original vision encoder, the time-series encoder is not pretrained. Therefore, we leave the pretraining of the time-series encoder as future work. It will be interesting to see whether the proposed dataset can enable large multimodal models to improve forecasting or classification accuracies, since the generated dataset allows them to associate the raw time series vector with common statistical concepts in the form of natural languages.

In terms of the dataset itself, our workflow for generating trend descriptions sheds the light on how descriptions regarding other time series properties can be generated, e.g., the change in volatility, or outlier identification using the extracted residuals. A more challenging task will be to generate descriptions for time series with multiple features, such as by studying their cross-correlations \citep{bracewell1965}.

\bibliographystyle{unsrt}
\bibliography{ref}
\newpage
\appendix
\section{Trend Dataset Details}\label{sec: dataset-details}
The 20 datasets involved in generating the TS-Insights dataset are listed below. 

\begin{table}[h]
\begin{tabular}{l|l|l} \hline  
\textbf{Dataset Name}& \textbf{Granularity} & \multicolumn{1}{|l}{\textbf{Number of Samples}} \\ \hline  
saugeenday\_dataset                                      & daily                & 201                                                              \\ \hline  
rideshare\_dataset\_without\_missing\_values             & hourly               & 1001                                                             \\ \hline  
pedestrian\_counts\_dataset                              & hourly               & 752                                                              \\ \hline  
oikolab\_weather\_dataset                                & hourly               & 1141                                                             \\ \hline  
nn5\_daily\_dataset\_without\_missing\_values            & daily                & 301                                                              \\
                                                         & tridaily             & 51                                                               \\ \hline  
                                                         & weekly               & 51                                                               \\ \hline  
m1\_yearly\_dataset                                      & yearly               & 100                                                              \\ \hline  
m1\_quarterly\_dataset                                   & quarterly            & 121                                                              \\ \hline  
m1\_monthly\_dataset                                     & monthly              & 351                                                              \\ \hline  
london\_smart\_meters\_dataset\_without\_missing\_values & half-hourly          & 1000                                                             \\ \hline  
kdd\_cup\_2018\_dataset\_without\_missing\_values        & hourly               & 800                                                              \\ \hline  
kaggle\_web\_traffic\_weekly\_dataset                    & weekly               & 800                                                              \\ \hline  
kaggle\_web\_traffic\_dataset\_without\_missing\_values  & daily                & 800                                                              \\ \hline  
hospital\_dataset                                        & monthly              & 500                                                              \\ \hline  
fred\_md\_dataset                                        & monthly              & 201                                                              \\ \hline  
elecdemand\_dataset                                      & half-hourly          & 102                                                              \\
                                                         & hourly               & 102                                                              \\  
                                                         & two-hourly           & 80                                                               \\  
                                                         & three-hourly         & 76                                                               \\   
                                                         & four-hourly          & 72                                                               \\   
                                                         & six-hourly           & 64                                                               \\   
                                                         & eight-hourly         & 17                                                               \\   
                                                         & twice-daily          & 17                                                               \\   
                                                         & daily                & 9                                                                \\ \hline  
covid\_mobility\_dataset\_without\_missing\_values       & daily                & 318                                                              \\ \hline  
covid\_deaths\_dataset                                   & daily                & 280                                                              \\ \hline  
cif\_2016\_dataset                                       & monthly              & 76                                                               \\ \hline  
bitcoin\_dataset\_without\_missing\_values               & daily                & 376                                                              \\ \hline  
australian\_electricity\_demand\_dataset                 & half-hourly          & 600                                                              \\ \hline  
                                                         & \textbf{Total}                &10360                                   
 \\ \hline \end{tabular}
\end{table}

The following augmentations maybe applied to a given time-series window each with a probability of $50\%$:
\begin{itemize}
    \item Jittering: Adding a Gaussian noise to the original time series, where the standard deviation of the Gaussian noise is set to be the standard deviation from a local rolling window of size 4.
    \item Scaling: Multiplying the original time series with a constant.
    \item Shifting: Adding a constant to the original time series.
    \item Smoothing: Convolving the original time series window with an average kernel of a randomly sampled size.
    \item Downsampling: Only keeping every other $k$ steps, where $k$ is another randomly sampled integer.
\end{itemize}
Note that multiple augmentations can be applied to get the final augmented window.

The holdout datasets are Electricity Demand (hourly, three-hourly, six-hourly, weekly), M3 (monthly, quarterly, other), M4 (hourly, daily, weekly, monthly, quarterly), Traffic (hourly, bi-hourly, four-hourly), and Weather (daily). 

On \href{https://huggingface.co/datasets/zhykoties/time-series-language-alignment}{HuggingFace}, we release ten window samples for each holdout and test dataset. However, due to the lack of human resources, the evaluation reported in Section \ref{sec: experiments} was done only using the first three samples from each dataset. Unfortunately, we only saved the original time series windows but not the augmented windows, but they should be able to be easily generated again using the scripted provided.

\section{Case Studies}\label{sec: case-studies}
\begin{figure*}[htbp]
    \centering
    \begin{subfigure}[b]{\textwidth}  
        \includegraphics[width=\textwidth]{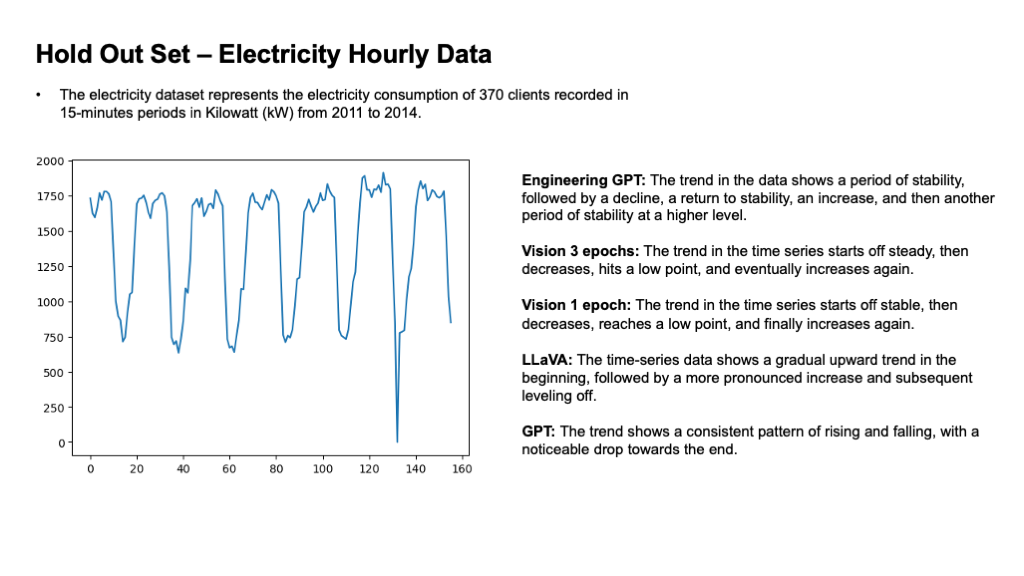}
    \end{subfigure}

    \begin{subfigure}[b]{\textwidth}  
        \includegraphics[width=\textwidth]{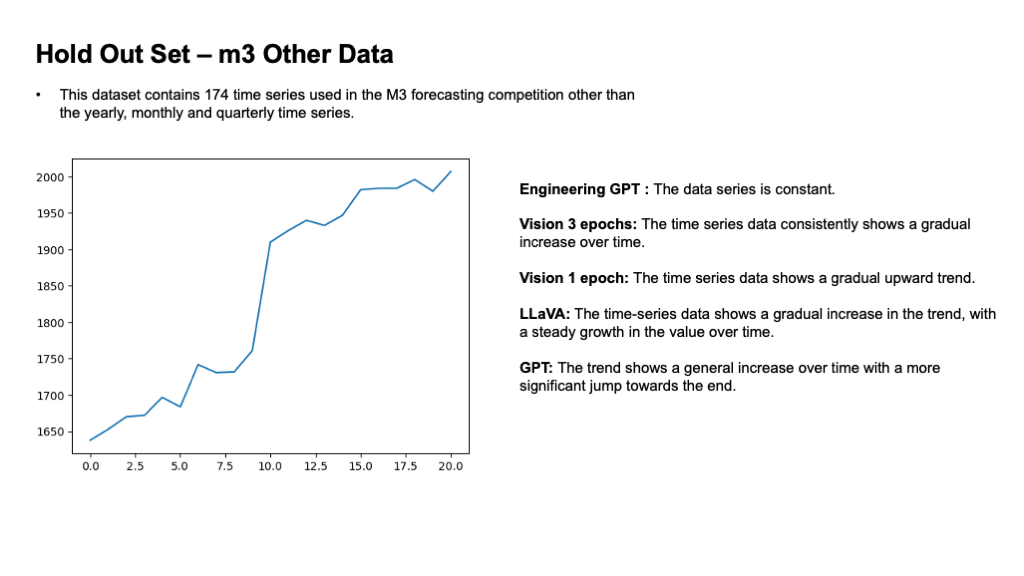}
    \end{subfigure}

\end{figure*}

\begin{figure*}[htbp]
    \centering
    \begin{subfigure}[b]{\textwidth}  
        \includegraphics[width=\textwidth]{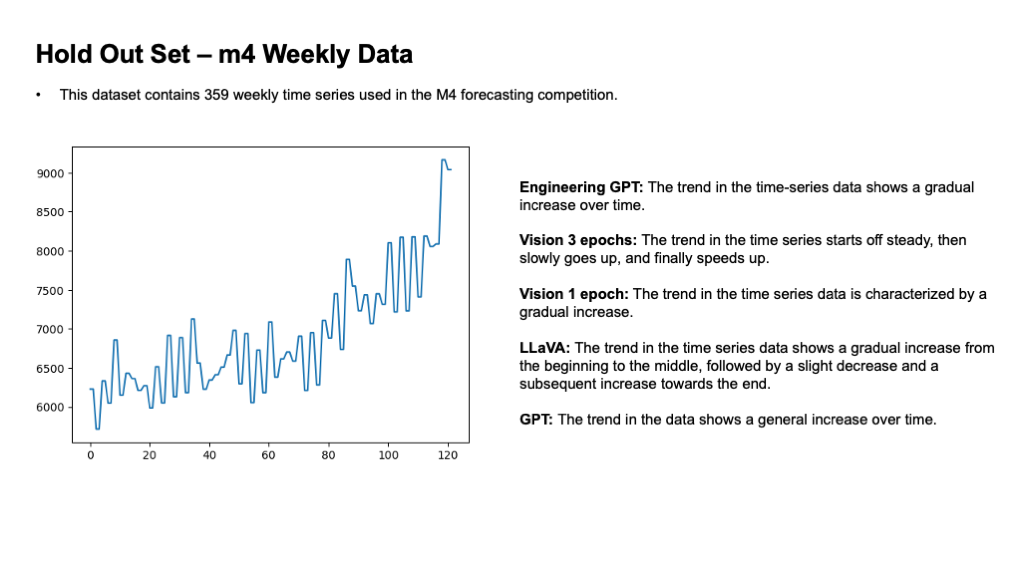}
    \end{subfigure}
    \begin{subfigure}[b]{\textwidth}  
        \includegraphics[width=\textwidth]{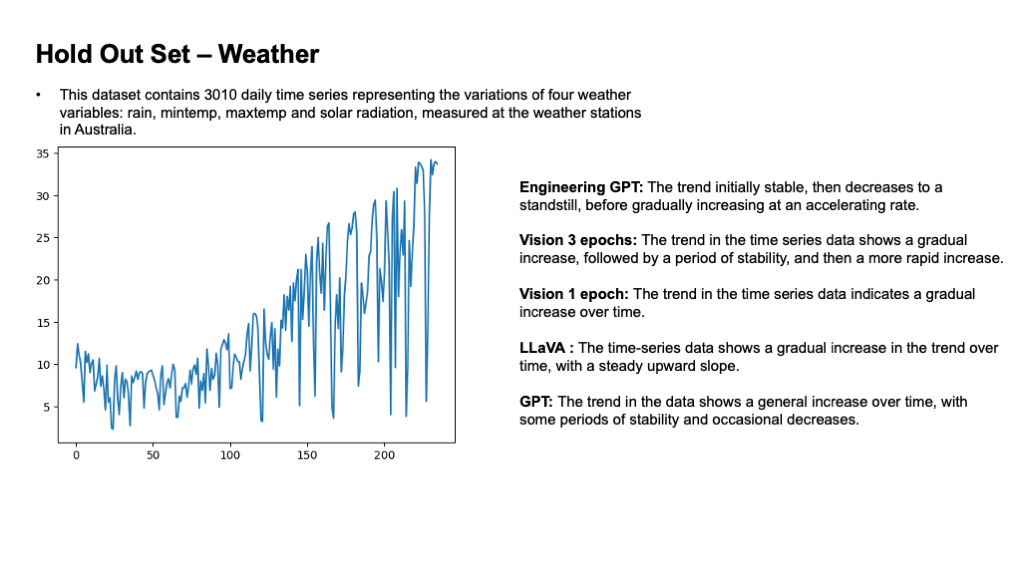}
    \end{subfigure}

    \begin{subfigure}[b]{\textwidth}  
        \includegraphics[width=\textwidth]{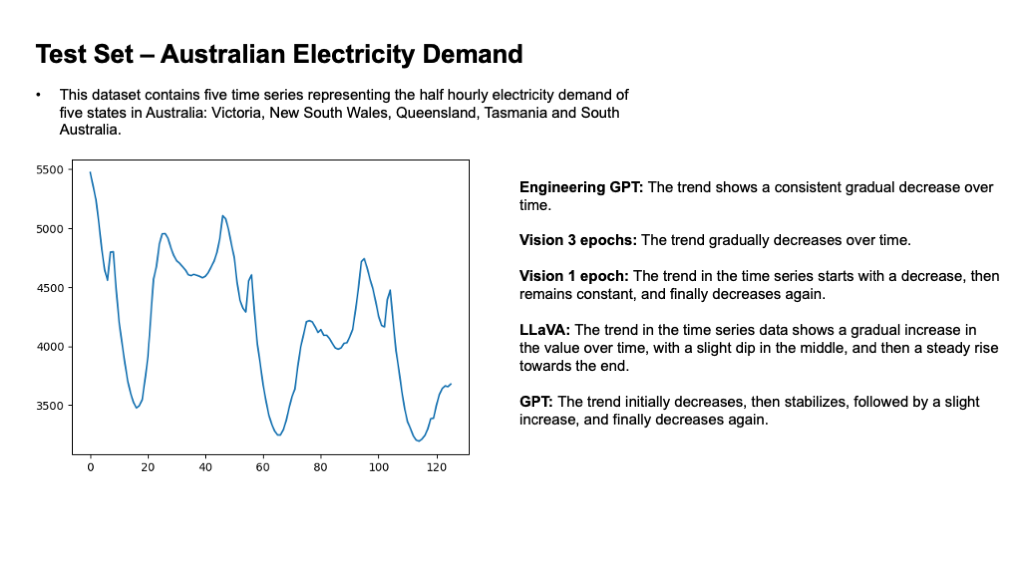}
    \end{subfigure}

\end{figure*}

\begin{figure*}[htbp]
    \centering
    \begin{subfigure}[b]{\textwidth}  
        \includegraphics[width=\textwidth]{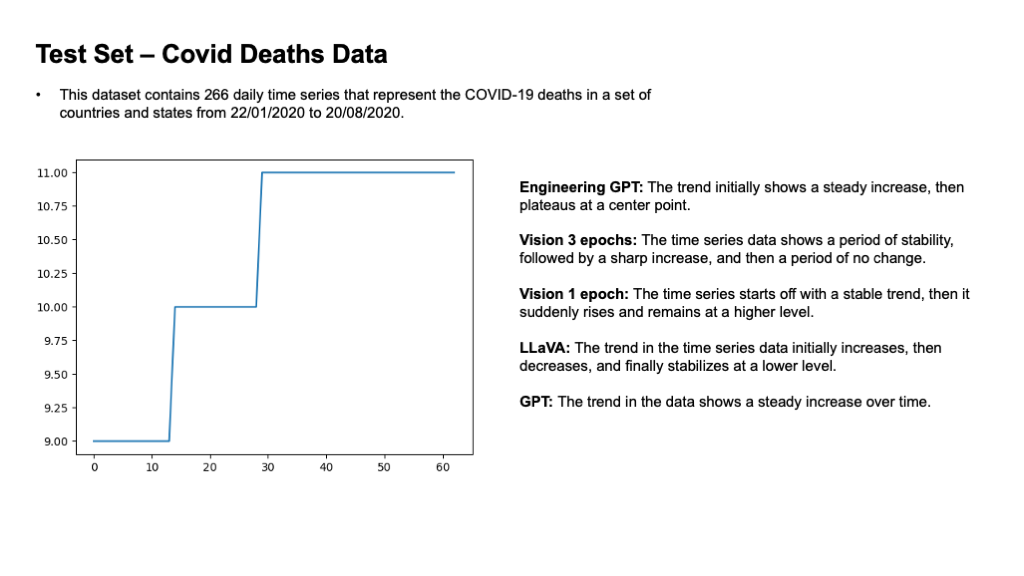}
    \end{subfigure}
    \begin{subfigure}[b]{\textwidth}  
        \includegraphics[width=\textwidth]{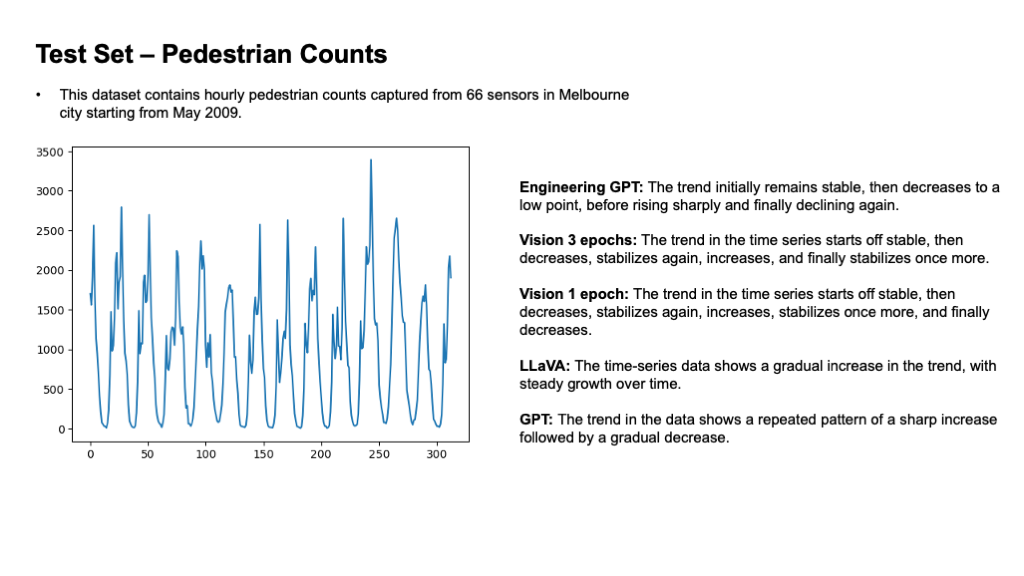}
    \end{subfigure}

    \begin{subfigure}[b]{\textwidth}  
        \includegraphics[width=\textwidth]{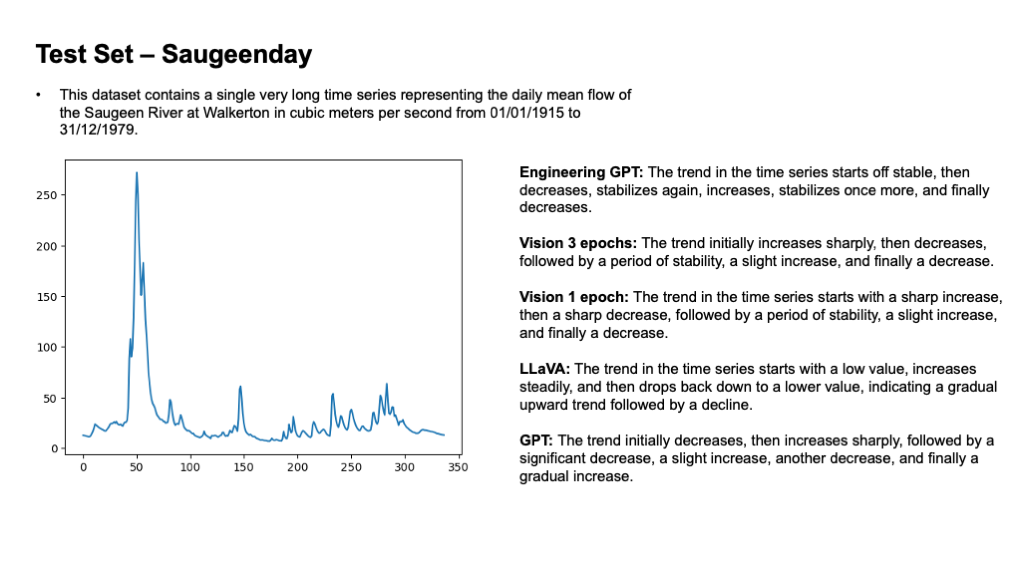}
    \end{subfigure}

\end{figure*}


\end{document}